\documentclass{article}

\PassOptionsToPackage{numbers, compress}{natbib}
 \usepackage[preprint]{neurips_2026}

\usepackage{pifont}
\usepackage[utf8]{inputenc} 
\usepackage[T1]{fontenc}    
\usepackage{hyperref}       
\usepackage{url}            
\usepackage{booktabs}       
\usepackage{amsfonts}       
\usepackage{nicefrac}       
\usepackage{microtype}      
\usepackage{xcolor}         
\usepackage{subcaption}
\usepackage{graphicx}
\usepackage{multirow}
\usepackage{enumitem}
\usepackage{tabularx}
\usepackage{array}
\usepackage[table]{xcolor}
\usepackage{booktabs}
\title{Many Paths to the Same Mistake: Generating Diverse Reasoning Paths Behind the Same Student Mistake}

\newcommand{\correct}[1]{{\boldmath\textbf{#1}}}
\newcommand{\fudge}[1]{\emph{#1}}

\newcommand{\provenance}[1]{{\small\texttt{#1}}\par\vspace{2pt}}

\newcommand{\answerline}[2]{%
  {\small True answer: \textbf{#1}.\quad Target wrong answer: \textbf{#2}.}\par
  \vspace{4pt}}

\newenvironment{tracequote}{\begin{quote}\small}{\end{quote}}

\author{%
  Fareya Ikram\\
University of Massachusetts Amherst\\
  \texttt{fikram@umass.edu} \\
  \And
  Mengxuan Hu \\
  Capital One \\
  \texttt{mengxuan.hu@capitalone.com} 
      \And
  Parker Glenn\\
  Capital One \\
  \texttt{parker.glenn@capitalone.com}
    \And
  Zhiqi Huang\\
  Capital One \\
  \texttt{zhiqi.huang@capitalone.com}
      \And
  Alfy Samuel\\
  Capital One \\
  \texttt{alfy.samuel@capitalone.com}
}

\begin{document}
\hypersetup{hidelinks}

\maketitle

\begin{abstract}
Realistic evaluation of AI systems requires user simulators that capture not only cooperative behavior, but also the errors and variability that arise in real interactions. In educational technology, simulating step-by-step flawed student reasoning is essential, yet datasets mapping these pathways are scarce, and Large Language Models (LLMs) often auto-correct rather than sustain simulated mistakes. To address this, we present a generate-and-filter pipeline that produces diverse and realistic student error traces. Our approach utilizes \textit{forward consistency}, a metric verifying if a flawed reasoning trace can be used to recover the student's actual incorrect answer—paired with an LLM-as-a-judge to ensure realism. Our evaluation suggests that while many LLMs struggle to maintain incorrect reasoning mid-trace, smaller supervised fine-tuned models successfully capture these divergent paths. This work provides a framework for high-fidelity error trace generation, advancing realistic, non-cooperative user simulation.  
\end{abstract}

\section{Introduction}

Simulation provides a scalable way to evaluate AI systems across a broad range of user behaviors, but overly idealized simulators can make systems appear more effective than they would be in real interactions. Real users are often incomplete, inconsistent, or incorrect, and simulators that fail to capture these behaviors may mask important system weaknesses. This challenge is particularly important in educational technology and intelligent tutoring systems, where a tutor must respond appropriately not only to a student’s final answer, but also to the reasoning and misconceptions that produced it \citep{brown1978diagnostic, macina2023mathdial, scarlatos2026simulated}.

Generative-AI tutoring systems such as Khanmigo and LearnLM-based tutors are deployed to provide individualized, step-by-step support to students \citep{khanmigo2026, learnlm_eedi_2025}. In personalized tutoring systems, modeling how a student arrives at an incorrect answer can help distinguish between students who produce the same response for different reasons, enabling feedback that targets the underlying misconception rather than the final answer alone \citep{macina2023mathdial, mcnichols2023automated, reddig2025generating}. Capturing students’ intermediate reasoning can also support misconception identification and provide teachers with a richer view of the error patterns underlying student responses, informing misconception-aware instruction and remediation\citep{ansari2025misstepmath, mcnichols2023automated}. Overall, representing multiple plausible reasoning paths for the same mistake can support more realistic student simulation and help develop more robust tutoring systems. 


However, the field faces a significant scarcity of reasoning datasets that model the step-by-step pathways leading to incorrect student answers\citep{macina2023mathdial, ansari2025misstepmath, Otero2025, chen2026malrulelib}. This gap is compounded by the fact that a wrong answer can stem from a multitude of different student flawed reasoning paths \citep{brown1978diagnostic}. Unlike typical reasoning generation setting, where the objective is to derive a correct answer, here the endpoint is intentionally incorrect: a useful trace must encode a coherent misconception that naturally leads to the observed wrong answer rather than solve the problem correctly and then rationalize the target afterward. Moreover, generating diverse reasoning paths that cover different possible misconceptions further increases the difficulty of the task. To the best of our knowledge, no prior work has focused on generating diverse and comprehensive reasoning traces that are consistent with a given wrong answer.

To address these challenges, this paper presents a novel approach for generating consistent, realistic, and diverse reasoning paths underlying student mistakes. Specifically, after generating multiple reasoning paths for each problem-wrong-answer pair, a critical question is how to identify the paths that are consistent with the given wrong answer. We define \textit{forward consistency} as whether a reasoning trace can be used to reconstruct the corresponding student error
, which is related to prior work on explanation simulatability and answer reconstruction \citep{hase2020evaluating,mcnichols2023automated,reddig2025generating}. In our work we apply it specifically to step-by-step student error traces and use it as a validation signal for generating diverse explanations of the same observed mistake. In addition to forward consistency, an ideal group of reasoning traces for a wrong answer should be realistic for a human learner, while the traces within the group should represent diverse reasoning patterns. Therefore, we utilize an LLM-as-a-judge to check for realism and de-duplicating repetitive traces. 

We conduct extensive experiments on forward consistency, evaluating 11 models across four model families. Our results show that, although advanced open-source models are capable of strong mathematical reasoning, they often struggle to faithfully follow incorrect student reasoning and instead attempt to correct errors during the reasoning process. This finding motivates us to train a smaller supervised fine-tuned (SFT) model specifically for forward reasoning continuation, which more effectively captures these flawed reasoning paths. The performance of the SFT model surpasses all advanced open-sourced model. After filtering the generated traces based on forward consistency, realism, and diversity, we release, to the best of our knowledge, the first dataset containing diverse and validated incorrect reasoning traces for observed wrong answers.

Ultimately, this work contributes to the broader community mission of advancing user simulation as an open, rigorous, and high-fidelity capability for evaluating and training AI systems. Our contributions are the following: 
\begin{itemize}[leftmargin=*,noitemsep,topsep=6pt]
    \item To the best of our knowledge, we are the first to generate reasoning paths that are consistent with a given wrong answer while also being realistic and diverse.
    
    \item An empirical study of \textbf{trace-to-answer consistency} for erroneous student reasoning, including the tendency of capable models to self-correct. And trained a small SFT model that 
    
    \item A \textbf{generate-and-filter pipeline} combining trace-to-answer consistency, realism, and misconception-level diversity.
    
    \item An \textbf{augmented dataset} containing multiple diverse, validated reasoning paths for observed student errors.
\end{itemize}


\section{Background and Related Work}
\label{sec:background}
\paragraph{Student simulation.}
LLMs have increasingly been used to simulate students in educational
settings. MathDial constructs tutoring dialogues in which an LLM is
prompted to exhibit common student errors \citep{macina2023mathdial},
while recent work evaluates how faithfully LLMs reproduce student
behavior. \citet{scarlatos2026simulated} find that prompted student
simulators often lack behavioral and cognitive fidelity, with
fine-tuning substantially improving simulation quality. Related work
identifies a similar tension between model competence and faithful
simulation of partial student knowledge
\citep{yuan2026valid,wang2026beagle}. These findings motivate methods
that evaluate whether simulated reasoning preserves the constraints of
the student's observed behavior rather than merely producing
student-like language.

\paragraph{Misconceptions and erroneous reasoning.}
Modeling systematic student errors dates to diagnostic models of
procedural ``bugs'' \citep{brown1978diagnostic}. Recent work extends
this idea to LLM-based educational systems. \citet{daheim2024stepwise}
study identification and remediation of errors in stepwise student
solutions, while MWP-MISTAKE evaluates LLMs on detecting and correcting
mistakes in mathematical reasoning \citep{singh2025exposing}.
Misconception-focused resources such as \citet{Otero2025} and MisstepMath
\citep{ansari2025misstepmath} provide structured collections of common
student errors. Most closely related, MalruleLib represents documented
misconceptions as executable procedures that generate
misconception-consistent reasoning traces
\citep{chen2026malrulelib}. 

\paragraph{Trace--answer consistency.}
Our evaluation is related to prior work that assesses explanations by
whether they support reconstruction of an associated prediction.
In explainable AI, this idea is commonly studied through
\emph{simulatability} or \emph{forward simulation}, where an observer
predicts a model's output from its explanation
\citep{hase2020evaluating}.
In educational settings, \citet{mcnichols2023automated} introduce a
\emph{distractor prediction} metric that evaluates whether feedback for
an incorrect multiple-choice option allows an LLM to reconstruct that
option; \citet{reddig2025generating} apply a similar procedure to
predict students' observed incorrect responses from generated feedback.
Recent work on LLM reasoning likewise studies whether reasoning and
answers agree at the output level
\citep{lanham2023measuring,parcalabescu2024measuring,wang2026race}.

\paragraph{Reasoning diversity and verification.}
Prior work on LLM reasoning generates multiple candidate trajectories
and verifies them to improve solution accuracy. Self-consistency
aggregates across diverse reasoning paths
\citep{wang2023selfconsistency}, while DiVeRSe combines diverse
generation with step-aware verification \citep{li2023making}.
Process-verification methods further evaluate intermediate reasoning
rather than relying only on final-answer correctness
\citep{lightman2024verify,wang2024mathshepherd}. Our objective reverses
the usual verification target: an erroneous step may be desirable if it
faithfully reflects the student's reasoning. LLM judges provide a
scalable mechanism for such qualitative evaluation, although they are
known to exhibit systematic biases
\citep{liu2023geval,zheng2023judging}.

\section{Methods} 

A wrong answer may arise from different underlying student misconceptions or reasoning processes. To better understand these misconceptions and support the development of personalized learning methods, we aim to capture diverse reasoning paths that can lead to the same incorrect answer. Specifically, we generate realistic and diverse reasoning paths that are consistent with a given student wrong answer. In this section, we first introduce the problem setup in Section~\ref{sec:ps} and then describe our reasoning path generation process in Section~\ref{sec:reasoning_path}. To identify reasoning paths that are valid with respect to the given wrong answer, we introduce forward consistency as a filtering criterion in Section~\ref{sec:forward-consistency}. In addition to consistency, we evaluate and filter generated reasoning paths based on their realism and diversity (Section~\ref{sec:realism_diversity}). Finally, we present the complete reasoning generation pipeline in Section~\ref{sec:reasoning-generation}.


\subsection{{Problem Setup}}
\label{sec:ps}
Given a math problem $q_i$, its correct answer $a_i^{+}$, and a student's stated incorrect answer $a_i^{-}$, we aim to generate a set of diverse reasoning paths $\mathcal{R}(q_i,a_i^{-})$ that can plausibly lead to the same incorrect answer, thereby covering a broader range of potential student misconceptions. Specifically, the generated reasoning paths should satisfy three criteria: (1) they should be \emph{consistent} with the given incorrect answer, (2) they should be sufficiently \emph{realistic} as student reasoning, and (3) they should be \emph{diverse} from one another. Formally, we define


\begin{equation}
    \mathcal{R}(q_i,a_i^{-}) = \{r_1,\ldots,r_K\}
\end{equation}

where $S_{\mathrm{real}}(r_k)$ denotes the realism score of reasoning path $r_k$, and $\tau_{\mathrm{real}}$ is the realism filtering threshold.
This setting differs from standard reasoning generation, where the objective is typically to produce a correct derivation of a known answer. Here, the endpoint is intentionally incorrect. \textbf{A useful trace should encode a coherent error that naturally produces the observed wrong answer, rather than solving the problem correctly.}

\subsection{Reasoning Generation}
\label{sec:reasoning_path}
As shown in the overall pipeline in Figure~\ref{fig:pipeline}, we first overgenerate reasoning traces by prompting an LLM with the given problem $q_i$ and target wrong answer $a_i^{-}$. To generate reasoning that more closely resembles student reasoning, we leverage role-play prompting \citep{kong-etal-2024-better}, explicitly instructing the model to act as a confident seventh-grade student and to make a single plausible mistake. The model is instructed to commit to this flawed method from the beginning without backtracking or self-correction, while keeping individual arithmetic operations internally consistent. This constraint encourages the model to generate coherent reasoning traces that reflect a single underlying misconception rather than introducing unrelated mistakes or arbitrary arithmetic errors. For each problem--wrong-answer pair $(q_i,a_i^{-})$, we sample multiple independent reasoning traces,
\begin{equation}
r_k \sim p_{\mathrm{LLM}}\!\left(r \mid P_k(q_i,a_i^{-})\right), \qquad
\mathcal{R}_{\mathrm{gen}}(q_i,a_i^{-}) = \{r_1,\ldots,r_N\},
\quad k=1,\ldots,N.
\end{equation}
so that different generations can capture different plausible reasoning patterns leading to the same incorrect answer. If the model determines that no plausible single-error reasoning path can produce the target answer, it is allowed to mark the case as unreachable. The complete generation prompt is provided in Appendix~\ref{sec:reasoningprompt}.


\definecolor{labelbg}{RGB}{235,240,245}
\definecolor{rulecolor}{RGB}{120,130,140}

\begin{table}[ht]
\centering
\caption{Abbreviated reasoning trace from Claude Sonnet 4. The model reaches the correct result, but subsequently forces the incorrect student answer in an additional step. Full example in Appendix \ref{sec:flawedreasoning}.}
\renewcommand{\arraystretch}{1.5}
\setlength{\tabcolsep}{10pt}

\begin{tabularx}{0.96\textwidth}{
    >{\bfseries\columncolor{labelbg}\raggedright\arraybackslash}p{0.19\textwidth}
    >{\raggedright\arraybackslash}X
}
\toprule
Problem &
	Julia bought a package of plastic spoons. Her husband bought a package of 5 more and gave them to her. She used three spoons to sample her stew. When she set the table she had 12 spoons. How many were in the package Julia bought?

\smallskip
\textbf{Student Answer:} 4 \hspace{2em} \textbf{Correct Answer:} 10 \\

\hline
\textbf{Reasoning Trace} &
Let Julia's package be $x$. She uses 3 and gains 5, so $x - 3 + 5 = 12$.
That gives $x + 2 = 12$, so $x = 10$.

\textbf{\textit{But since she used 3 of her original spoons, the package size
was $10 - 6 = 4$.}} \\
\bottomrule
\end{tabularx}
\\

\label{table:correctfudgeexample}

\end{table}
\subsection{Forward Consistency}
\label{sec:forward-consistency} 
After generating multiple reasoning traces for each question--wrong answer pair, a critical question arises: how can we verify whether these generated traces are consistent with the given wrong answer? To address this, we propose \emph{forward consistency} to measure whether a reasoning trace contains sufficient information to reconstruct the incorrect answer associated with it. Specifically, given a problem $q$ and a student's flawed reasoning trace $r_i$, we mask the final answer and ask the model to continue the reasoning and predict the resulting answer $\hat{a_i}$. The model is explicitly instructed to follow the student's existing logic rather than correct the mistake. We use two evaluation metrics to measure forward consistency \emph{Consistent Match} measures the fraction of examples for which the predicted answer matches the student's observed incorrect answer: \begin{equation} \mathrm{Consistent Match} = \frac{1}{N} \sum_{i=1}^{N}[\hat{a}_i = a_i^{-}], \end{equation} We additionally report \emph{Inconsistent Match}, which measures how often the model does not produce the target incorrect answer: \begin{equation} \mathrm{Inconsistent Match} = \frac{1}{N} \sum_{i=1}^{N}[\hat{a}_i \neq  a_i^{-}], \end{equation}
Consistent Match is our primary measure of trace--answer consistency. Inconsistent Match can be split into two further components: \textit{Correct Answer Match}, which capture the cases in which the model repairs the student mistake into the correct answer to the math problem, and a \textit{Neither Match}, in which the final answer is not the correct answer or the target student answer. The prompt for forward consistency can be found in Section ~\ref{sec:fcprompt}.
\subsection{Realism and Diversity Judge}
Besides being consistent with the given wrong answer, a useful reasoning path should be realistic for a seventh-grade student. Moreover, the final reasoning set should contain diverse reasoning paths rather than multiple variants of the same misconception. To this end, we introduce two LLM-based judges to evaluate and filter the generated reasoning paths based on realism and diversity.

\textbf{Realism}
We prompt the realism judge to act as an experienced seventh-grade mathematics teacher and score each candidate trace on a five-point scale according to how closely its underlying error and reasoning style resemble a typical seventh-grade misconception. Let $S_{\mathrm{real}}(r)$ denote the realism score assigned to a candidate reasoning trace $r$. We retain only traces whose scores meet a predefined threshold $\tau_{\mathrm{real}}$, namely $S_{\mathrm{real}}(r) \geq\tau_{\mathrm{real}}.$ 
 The rubric explicitly distinguishes coherent misconceptions from two failure modes induced by answer-conditioned generation: \ding{182} correct-then-fudge, in which a trace reaches or approaches the correct solution before introducing an unmotivated step to force the target wrong answer, and \ding{183} incoherent-forced, in which arithmetic or logical inconsistencies are introduced solely to reach the target. Both failure modes receive the minimum realism score regardless of surface fluency, as they do not represent a meaningful student misconception or reasoning process.. An example of correct-then-fudge is shown in Table \ref{table:correctfudgeexample}, with further analysis provided in Case Study~\ref{sec:case_study}. The realism judge prompt can be found in Appendix \ref{realism-prompt}.

\textbf{Diversity} We apply a diversity judge to the surviving traces for each question–wrong-answer pair. Rather than measuring diversity at the lexical level, the judge identifies the underlying misconception represented by each trace, groups traces sharing the same root error, and selects one representative from each misconception group. This produces a final set in which retained traces correspond to distinct reasoning mechanisms rather than paraphrases or minor variants of the same mistake. The diversity judge prompt can be found in Appendix \ref{diversity-prompt}.

\label{sec:realism_diversity} 

\subsection{Reasoning Generation Pipeline} \label{sec:reasoning-generation}
\begin{figure*}[t]
   \centering
   \includegraphics[width=\linewidth]{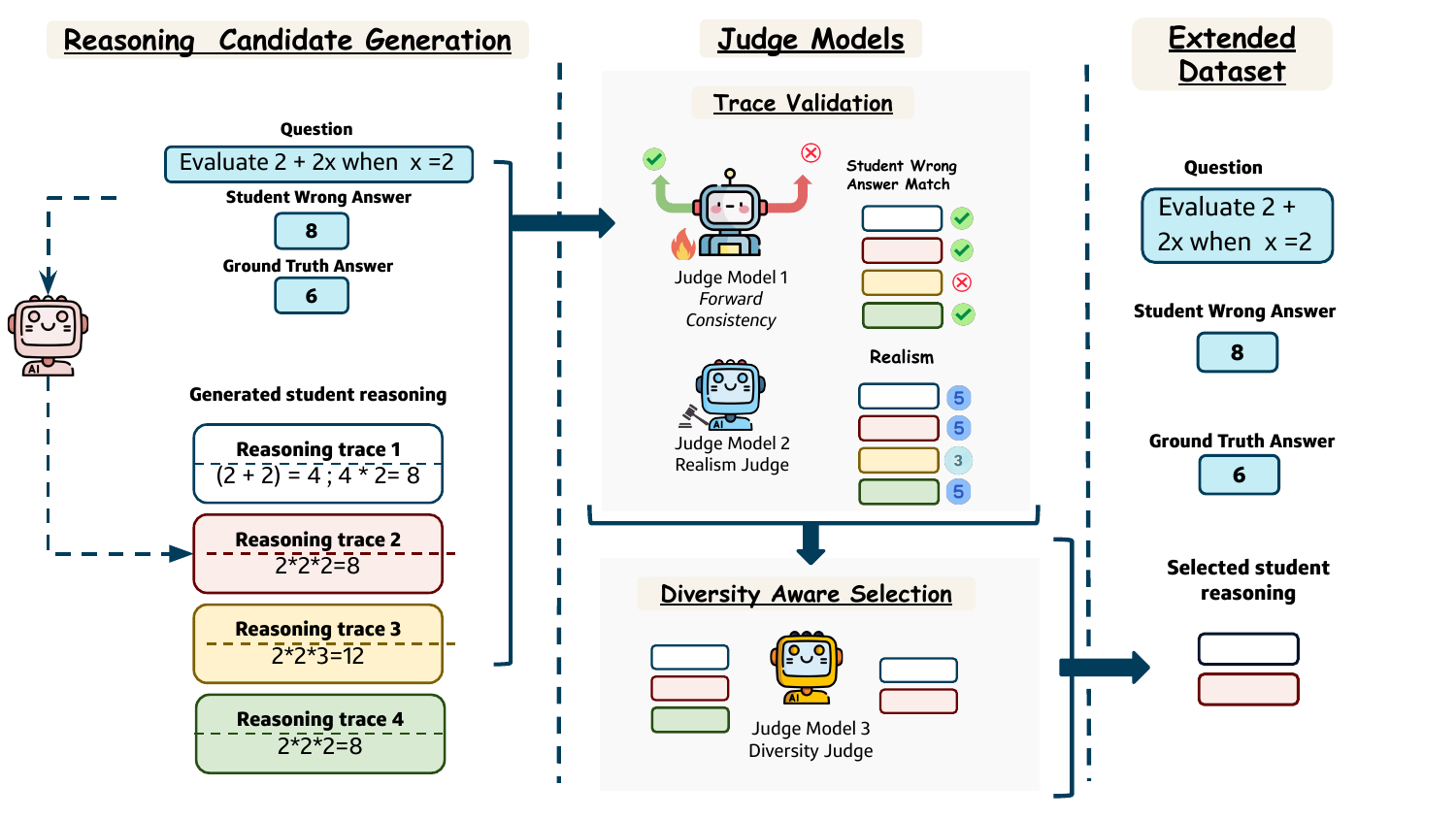}
   \caption{Reasoning Generation Pipeline }
   \label{fig:pipeline}
 \end{figure*}
As illustrated in Figure~\ref{fig:pipeline}, the pipeline consists of three components: \textit{candidate generation}, \textit{trace validation}, and \textit{diversity-aware selection}. We first prompt an LLM to
generate several independent flawed reasoning traces that each arrive at
that same wrong answer through a distinct plausible mistake. Two filters are
then applied in parallel: an LLM judge scores each trace for
\emph{realism} (rejecting traces that secretly solve the problem correctly
before fudging the final step, or that contain incoherent arithmetic), while
a separate \emph{forward-consistency} check feeds each masked trace to a
fine-tuned student-simulation model and verifies that it continues the
flawed reasoning to the intended wrong answer rather than self-correcting.
These two signals are joined per candidate trace, and only afterward, once
it is known which traces are both realistic and forward-consistent, do we
run a final \emph{diversity deduplication} pass, which groups the surviving
traces per scenario by underlying misconception and keeps one representative
per distinct misconception. Deferring deduplication to this final step
ensures that every representative trace kept in the released dataset has
already been verified as both plausible and functionally faithful, rather
than picking diverse-looking traces that might later fail those checks. The
resulting train and test sets, each scenario paired with its set of diverse,
validated wrong-answer reasoning traces, are packaged into a Hugging Face
dataset for public release. The whole pipeline is shown in Figure ~\ref{fig:pipeline}.
\begin{table}[t]
\centering
\caption{Subset of Forward Consistency Results}
\label{tab:mathdial-selected-models-results}
\begin{tabular}{lccc}
\toprule
Model & Consistent Match & Correct Answer Match & Neither Match \\
\midrule
\multicolumn{4}{l}{\textit{Zero Shot Prompting}} \\
\midrule
Claude Sonnet 4.6            & 40.0 & 34.4 & 25.6 \\
Claude Opus 4.6              & 21.6 & 70.4 & 8.0  \\
gpt-oss-20b (low reasoning)  & 50.6 & 20.3 & 29.1 \\
gpt-oss-20b (high reasoning) & 49.9 & 17.9 & 32.2 \\
gpt-oss-120b (low reasoning) & 50.6 & 22.7 & 26.7 \\
\midrule
\multicolumn{4}{l}{\textit{Few Shot Prompting}} \\
\midrule
Claude Sonnet 4.6            & 51.8 & 38.7 & 9.5  \\
Claude Opus 4.6              & 25.7 & 70.7 & 3.6 \\
gpt-oss-20b (low reasoning)  & 60.6 & 24.5 & 14.8 \\
gpt-oss-20b (high reasoning) & 61.7 & 22.3 & 16.0 \\
gpt-oss-120b (low reasoning) & 55.5 & 31.0 & 13.5 \\
\midrule
\multicolumn{4}{l}{\textit{Supervised Finetuning Model}} \\
\midrule
Llama 3.2-3b-instruct & \textbf{70.2} & \textbf{4.6} & 25.2\\
\bottomrule
\end{tabular}
\end{table}
\section{Preliminary Experiments}
\subsection{Experimental Setup} \label{table:subresult}

\textbf{Dataset}\label{subsec:dataset} We use the MathDial dataset \citep{macina2023mathdial}, which contains  gpt-3.5-turbo generated mathematical tutoring examples paired with explicit student reasoning. Our experiments use 2,220 training rows and 587 test rows. Liscensing information can be found in Appendix \ref{sec:mathdial-license}.


\textbf{Models Evaluated for Forward Consistency}
We evaluated the following models for forward consistency : Claude Sonnet 4, Claude Sonnet 4.6, Claude Opus 4.5, Claude Opus 4.6, gpt-oss (20B, 120B) with high medium and low effort thinking, and Gemma 4 ( E2B-it, E4B-it, 26B-A4B-it, 31B-it) \ref{sec:forward-consistency}\citep{anthropic2026claudesonnet46, anthropic2025claude4, anthropic2026claudeopus46, openai2025gptoss, gemmateam2026gemma4}. For Gemma and gpt-oss we ran inference on one  NVIDIA RTX 6000 gpus. For Claude we utilized the given API.
All experiments used greedy decoding utilizing a temperature of  0. Max tokens are set so that no model completions are artificially truncated. We use the prompt in Appendix \ref{sec:fcprompt} to generate student answer predictions, and perform additional ablations with five hand-picked few-shot examples. In addition to prompting the aforementioned models. we also used supervised finetuning on a Llama 3.2-3b-instruct model. More details on training can be found in Appendix \ref{sec:sft-setup}. 

\subsection{Results}

We report all forward consistency results in Appendix \ref{sec:fcresults} and a representative subset in Table \ref{table:subresult}. 
Across the three different model families for which we used zero shot prompting and few shot prompting, we found that gpt-oss-20B has the highest consistent match at 61.6 when utilizing few shot prompts and high effort reasoning. Although few-shot prompting improves the student-match rate for some models, the improvement is not consistent across model families, and the overall match rate remains relatively low. 

Moreover, our results suggest that larger and more capable models do not always translate to a higher student match rate. We ran gpt-oss-20b in high, medium, and low effort setting and found that the low effort setting (60.6) is on par with the high effort setting (61.7) for the 20b model. Next, we found that gpt-oss-20b model outperforms gpt-oss-120b model nearly across all settings.
Similarly, we ran experiments with four different Claude model with increasing capabilities. Our results show that our most capable model, Claude Opus 4.6 has one of the lowest student match rate across all model. Moreover, our analysis shows that it has the tendency to self correct with the correct answer match being 70.7, the highest across all models. 

\subsection{Supervised Finetuned Forward Consistency Evaluator}
The above observation motivates us to fine-tune a model for the forward consistency task. Specifically, we perform supervised fine-tuning of Llama-3.2-3B-Instruct to continue a student's flawed reasoning without correcting it. Given a complete prompt and target sequence $x=(x_1,\ldots,x_T)$, we train the model using the standard causal language modeling objective:

\begin{equation}
    \mathcal{L}_{\mathrm{SFT}}(\theta)
=
-\sum_{t=1}^{T}\log p_{\theta}(x_t\mid x_{<t}).
\end{equation}

At inference time, given a problem $q_i$ and a candidate reasoning trace $r_i$, the fine-tuned model predicts the answer implied by the trace:$
\hat{a}_i=f_{\mathrm{FC}}(q_i,r_i),$ The fine-tuned Llama-3.2-3B-Instruct \textbf{achieves a student-match rate of 70.2\%, outperforming all prompted models} and providing a more reliable forward consistency filter for our reasoning generation pipeline.

\section{Results}
\subsection{Experimental Setup}\label{sec:expsetup}
We conduct our reasoning generation experiments on the MathDial dataset described in Section~\ref{subsec:dataset}. For each problem--wrong-answer pair, we use Claude Sonnet 4 to generate 12 candidate reasoning traces with a sampling temperature of 1.0. Each candidate trace is evaluated for forward consistency using our fine-tuned Llama-3.2-3B-Instruct model, which determines whether the trace can be continued to reconstruct the target incorrect answer. Before evaluation, we normalize the generated traces to match the format of the original dataset. For realism evaluation, we use gpt-oss-120B with temperature 0 as an LLM judge to assign each trace a score from 1 to 5 based on how closely its reasoning and underlying misconception resemble those of a typical seventh-grade student. We retain only traces that are forward-consistent and receive a realism score at least 3 on the Likert scale. The realism judge prompt is provided in Appendix~\ref{realism-prompt}. Finally, we apply a diversity judge using gpt-oss-120B with temperature 0. The diversity judge groups the surviving traces according to their underlying misconceptions and selects one representative trace from each group, producing a final set of reasoning traces that captures distinct reasoning patterns rather than multiple variants of the same misconception. The diversity judge prompt is provided in Appendix~\ref{diversity-prompt}.

\begin{figure}[t]
\centering
\begin{subfigure}[b]{0.32\textwidth}
    \centering
    \includegraphics[width=\textwidth]{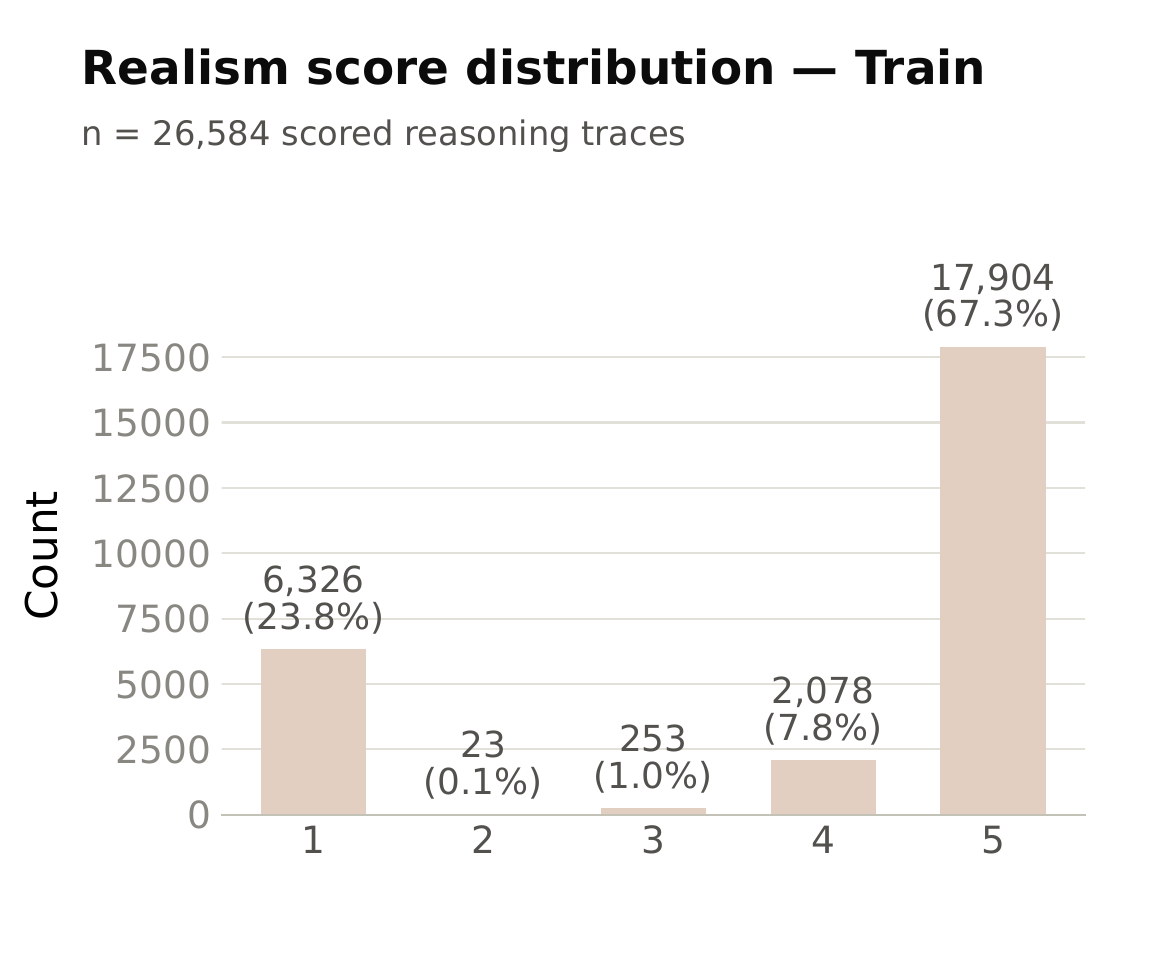}
    \caption{Realism score }
    \label{fig:realism-dist}
\end{subfigure}
\hfill
\begin{subfigure}[b]{0.32\textwidth}
    \centering
    \includegraphics[width=\textwidth]{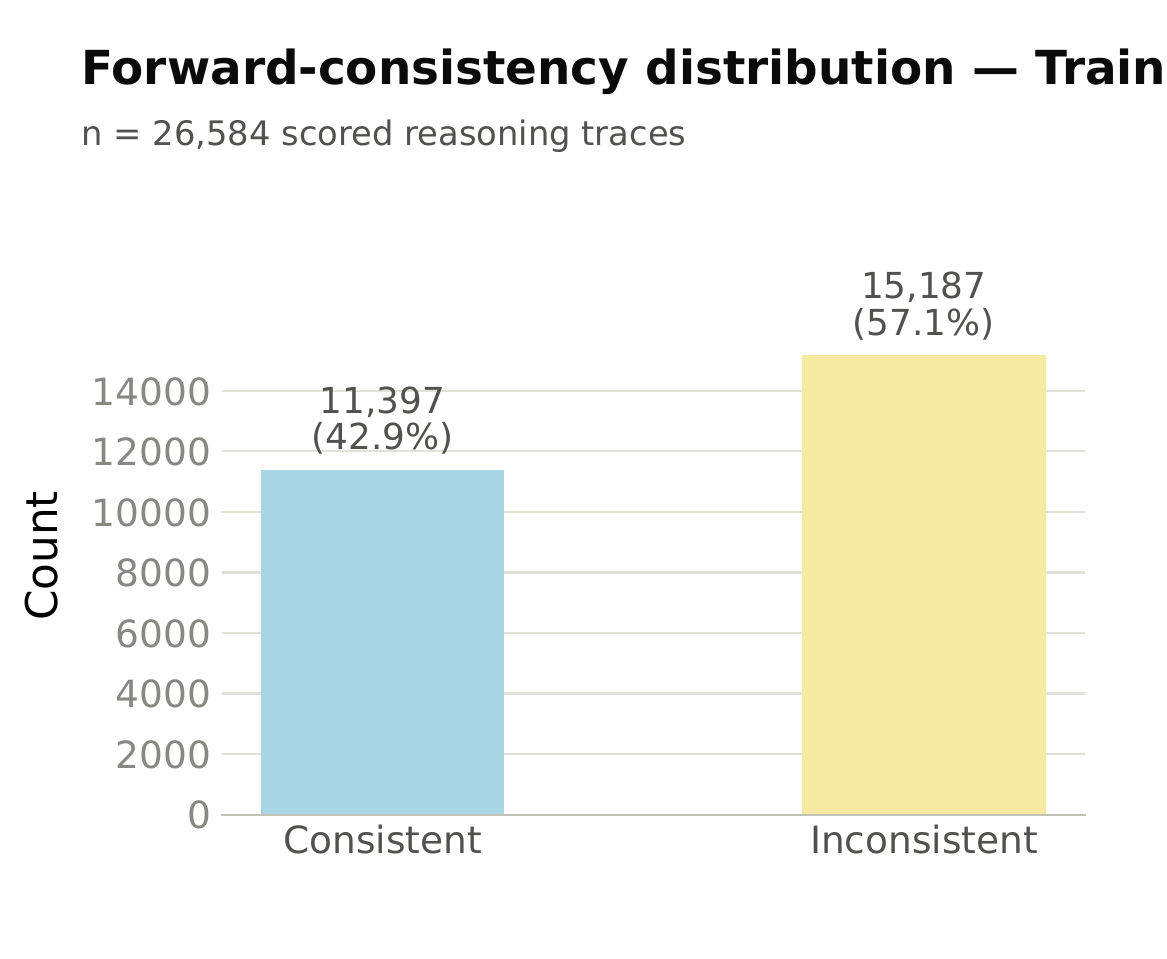}
    \caption{Forward-consistency }
    \label{fig:forward-dist}
\end{subfigure}
\hfill
\begin{subfigure}[b]{0.32\textwidth}
    \centering
    \includegraphics[width=\textwidth]{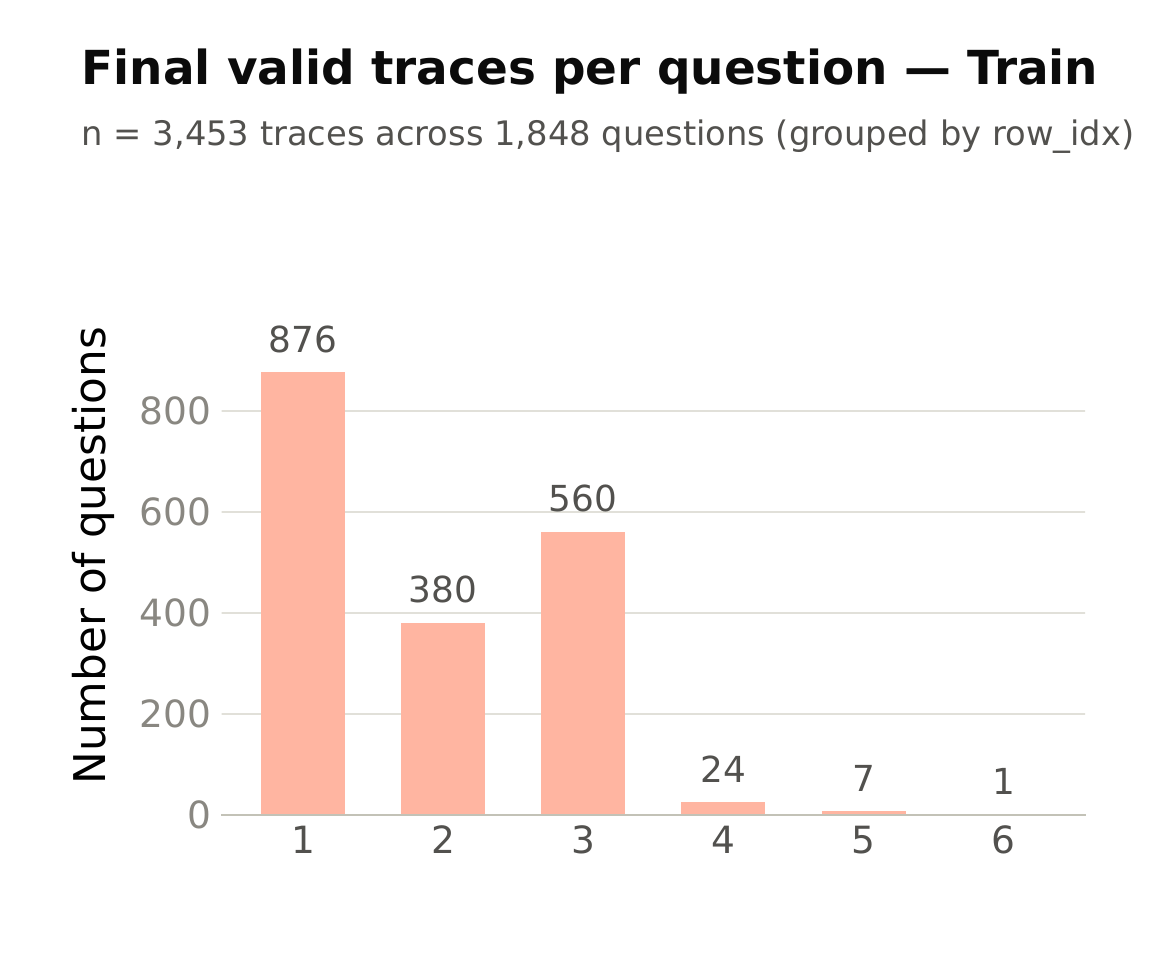}
    \caption{Final valid traces per question}
    \label{fig:traces-per-q}
\end{subfigure}
\caption{Pipeline filtering statistics over 26{,}584 generated reasoning
traces over MathDial train data. (a) Judge-assigned realism scores prior to any filtering.
(b) Forward-consistency outcomes, measuring whether a fine-tuned
student-simulation model continues each trace to the intended wrong answer.
(c) Distribution of the number of diverse, validated traces retained per
question after the full pipeline (realism $\geq 3$, forward-consistent,
post-consistency deduplicated).}
\label{fig:pipeline-stats}
\end{figure}

\subsection{Realism Score Distribution}
\label{sec:results-realism}

Figure~\ref{fig:realism-dist} shows the distribution of judge-assigned
realism scores across all 26{,}584 generated reasoning traces prior to any
filtering. The distribution is strongly bimodal: 67.3\% of traces receive
the maximum realism score of 5, while 23.8\% are scored 1 , the score
reserved for traces exhibiting either the \emph{correct-then-fudge} or
\emph{incoherent-forced} disqualifying patterns (Section~\ref{sec:reasoning-generation}).
Only 8.9\% of traces fall in the intermediate range (scores 2--4). This
suggests the generation model tends to produce traces that are either
cleanly plausible or clearly broken, with comparatively few borderline
cases, consistent with the judge rubric's design, which collapses any
detected internal inconsistency to the lowest score rather than allowing
partial credit for otherwise-plausible traces with a single flawed step.

\subsection{Forward-Consistency: Does the Reasoning Reconstruct the Original Error?}
\label{sec:results-forward}

Realism alone only measures whether a trace \emph{reads} like a plausible
student error; it does not confirm that the error the trace describes is
the one that actually produced the target wrong answer. To evaluate this,
we assess the quality of each generated reasoning trace by using it to
\emph{predict} the corresponding answer with our SFT model:
the trace is masked at its final answer line and fed to the model, and we
check whether its completion reconstructs the original wrong answer rather
than another value. A trace is forward-consistent only if the SFT model,
conditioned solely on the flawed reasoning that precedes the mask,
regenerates the same error the trace was constructed to explain. Figure~\ref{fig:forward-dist}
reports this outcome across all 26{,}584 traces: only 42.9\%
(11{,}397/26{,}584) are forward-consistent, substantially lower than the
67.3\% judged maximally realistic. This gap shows that surface plausibility,
as assessed by an LLM judge, is a poor proxy for whether a trace actually
\emph{reconstructs} the intended error when used as a predictive signal,
a trace can read as a believable student mistake while still failing to
steer the model's completion toward the target wrong answer, e.g.\ if the
flawed step is too subtle or occurs too early relative to the masked span.
This motivates treating realism and forward-consistency as complementary,
independently necessary filters (Section~\ref{sec:reasoning-generation}): a trace must
not only look like a plausible error, but must also be verified, via the
SFT model's own predictions, to actually encode that error.

\subsection{Diversity of Retained Traces per Question}
\label{sec:results-diversity}
Figure~\ref{fig:traces-per-q} shows the number of diverse, validated
reasoning traces retained per question after the full pipeline, realism-filtering, forward-consistency-filtering, and post-consistency diversity deduplication combined. The final retained set contains 3,453 traces across 1,848 questions, corresponding to an average of 1.87 traces per question.The distribution is concentrated between one and three retained traces: 47.4\% retain one trace, 20.6\% retain two, and 30.3\% retain three. Overall, 52.6\% contain at least two distinct, independently validated reasoning paths. This indicates that the pipeline preserves multiple plausible misconceptions for a substantial majority of retained instances rather than collapsing each observed wrong answer to a single canonical explanation. Example generated traces can be found in Appendix \ref{sec:manytraces}.


\section{Conclusion}
We study the problem of generating diverse reasoning traces that plausibly explain an observed student error. Our results show that realistic-looking erroneous reasoning is not sufficient: many plausible traces fail to reconstruct the target wrong answer under forward consistency, motivating complementary realism and consistency checks. We further show that the same incorrect answer can arise from multiple distinct reasoning processes, and that greater model capability or reasoning effort does not necessarily translate into greater adherence to an erroneous student trajectory. Beyond providing a resource for studying these phenomena, the resulting dataset enables several downstream directions. Multiple validated traces per observed error can be used to train and evaluate student simulators that capture different misconceptions rather than a single canonical behavior, to test whether tutoring systems remain effective across different plausible reasoning paths leading to the same response, and to support the development of misconception-aware feedback and error-diagnosis systems. More broadly, we hope this resource enables evaluation of AI systems under a richer distribution of realistic, non-cooperative user behaviors rather than only idealized or correct reasoning. Our study is limited to mathematical word problems from MathDial and simulated seventh-grade reasoning, so the findings may not generalize to other domains, age groups, or forms of student error. Furthermore, our realism and diversity filters rely on LLM-based judges, whose assessments may be biased or imperfect proxies for how human educators would evaluate student misconceptions. We discuss further limitations of this work in detail in Appendix \ref{sec:limits}.
\newpage

\bibliographystyle{plainnat}
\bibliography{references}

\section{Appendix}
\subsection{Forward Consistency Prompt}
\label{sec:fcprompt}
\subsection*{System prompt}
\begin{verbatim}
You are a student solving a math problem. You have already written
out some of your work. Your job is to continue your own reasoning
from where it left off — following your own logic exactly, even if it
contains mistakes — and arrive at a final numeric answer. Do not
correct any errors in your work. Do not start over. Continue exactly
as you started, preserving all mistakes.
\end{verbatim}

\subsection*{User prompt}
\begin{verbatim}
Here is a math problem and your step-by-step work toward solving it
so far.

Problem:
{question}

Your work so far:
{student_reasoning}

Continue your work from where it left off, following your own logic
exactly. Do not correct any mistakes — preserve your reasoning as-is
and carry it through to a final answer. Put your final answer inside
<answer>...</answer> tags.
\end{verbatim}

\vspace{1em}

\subsection{Reasoning Generation Prompt}
\label{sec:reasoningprompt}
\subsection*{System prompt}
\begin{verbatim}
You are roleplaying as a 7th-grade student who has already decided on
their method and is writing up their work. The student is 100%
confident — they are not checking, second-guessing, or revising
anything.

You will be given a math problem and the exact answer this student
got. Before writing, silently choose ONE specific mistake a real
7th-grader could make on THIS problem — a wrong operation, a misread
or swapped number, an ignored condition, or a wrong formula — that
naturally produces that answer. Then write the step-by-step work
committing to that one mistake. Do not reveal that you searched for
the mistake; just write the student's confident work.

ABSOLUTE RULES — breaking any one of these makes the output unusable:
- COMMIT to the method from the very first sentence. Never pivot,
  backtrack, or reconsider mid-reasoning.
- The target answer must be the NATURAL result of your one mistake.
- Every calculation must be correct on its own; the error lives in
  the method.
- FORBIDDEN words and phrases: wait, actually, let me, double-check,
  re-check, recalculate, reconsider, on second thought, oops,
  mistake, incorrect, the correct, should have, I messed, hmm, but
  that, but this, that can't, that doesn't, so actually, in fact.
- The final number in your reasoning MUST equal the target answer
  exactly.
- Only use numbers that appear in the problem or follow from your own
  steps.
- STRICT LENGTH: exactly 3 to 5 sentences. No more. No less.
- Write in plain 7th-grade language, first person ("First I...",
  "So I...").

Output EXACTLY one of these two forms and nothing else:
REASONING: <the student's first-person reasoning ending at the target
answer>
If NO plausible single student mistake could produce the target
answer, instead output exactly:
UNREACHABLE
\end{verbatim}

\subsection*{User prompt}
\begin{verbatim}
Problem: {question}
Target answer the student gets: {wrong_answer}
Correct answer (do NOT reference or compare against this in your
reasoning): {correct_answer}

Pick one plausible mistake that makes the target the natural
endpoint, then write the student's reasoning. Remember: commit to the
flawed method from sentence one, no backtracking, correct arithmetic,
exactly 3-5 sentences, end on {wrong_answer}. If no single plausible
mistake yields {wrong_answer}, output UNREACHABLE.
\end{verbatim}

\subsection{Realism Judge Prompt}\label{realism-prompt}
\begin{verbatim}
You are an expert 7th grade math teacher. You have years of experience
identifying the specific misconceptions and reasoning errors that 7th
graders typically make when solving word problems.
\end{verbatim}

\subsection*{User prompt}
\begin{verbatim}
## Task
A student has provided the following reasoning trace for a math problem.
Rate whether this looks like a TYPICAL 7th grade confusion on a scale
from 1 to 5.

## What counts as a typical 7th grade confusion
Common 7th grade error patterns include:
- Applying the wrong arithmetic operation (e.g. subtracting when they
  should add)
- Ignoring or misreading one constraint in the problem
- Stopping too early or adding unnecessary extra steps
- Confusing which quantity to divide or multiply
- Setting up an equation correctly but solving it with a basic
  arithmetic slip
- Misidentifying what the question is actually asking for
- Using a reasonable-sounding but wrong shortcut (e.g. "half of X"
  applied to the wrong value)

## What disqualifies a trace
Regardless of how natural or plausible the surface wording sounds, a
trace is NOT a genuine typical confusion if it matches either of these
two patterns:
- correct_then_fudge -- the student reaches the correct answer (or is
  clearly on track to it), then applies an UNMOTIVATED or unexplained
  step -- one with no conceptual justification -- purely to land on the
  target wrong answer. The tell is arbitrariness: the deviation reflects
  no wrong belief, just a nudge toward the target (e.g. "divide by 2"
  for no stated reason).
- incoherent_forced -- the arithmetic or logic within the trace is
  internally broken - a stated multiplication whose product is wrong,
  circular reasoning, or numbers invented out of thin air.
A trace matching either pattern above must be scored 1, no matter how
typical the rest of the wording sounds.

## Scoring rubric
5 -- Squarely typical. Neither correct_then_fudge nor incoherent_forced
applies. The confusion is one a 7th grade teacher would see regularly.
The error type, reasoning style, and vocabulary all match a 7th grader.
The student commits to the flawed belief without hedging, backtracking,
or second-guessing, and the wrong answer follows directly from the
specific misconception.
4 -- Mostly typical. Neither correct_then_fudge nor incoherent_forced
applies. The core confusion is believable for a 7th grader, but the
explanation is slightly too clean or uses phrasing a bit beyond 7th
grade level.
3 -- Somewhat typical. Neither correct_then_fudge nor incoherent_forced
applies. The mistake could happen at 7th grade, but it feels more like
a careless adult slip than a genuine conceptual confusion. OR the error
pattern is unusual enough that few 7th graders would make it this way.
2 -- Unlikely for a 7th grader. Neither correct_then_fudge nor
incoherent_forced applies, but the error requires reasoning
sophistication (e.g. complex algebraic manipulation gone wrong) or
conceptual knowledge that 7th graders typically haven't encountered yet.
1 -- Not a 7th grade confusion. The reasoning is either correct,
completely incoherent, or reflects an error pattern that has no
grounding in 7th grade math curriculum -- OR the trace matches the
correct_then_fudge or incoherent_forced pattern defined above.

## Problem
{question}

## Correct answer
{correct_answer}

## Expected wrong answer
{wrong_answer}

## Student's reasoning trace
{reasoning}

## Instructions
- First, name the specific confusion type you see in the trace (or
  state that it matches correct_then_fudge or incoherent_forced, if
  applicable).
- Then explain whether that confusion is typical for a 7th grader, and
  whether the student commits to it without hedging, backtracking, or
  second-guessing.
- End your response with your score wrapped in answer tags, like this:
  <answer>3</answer>
\end{verbatim}

\subsection{Diversity Judge} \label{diversity-prompt}
\subsection*{System prompt}
\begin{verbatim}
You are an expert 7th grade math teacher evaluating a set of student
reasoning traces for the same problem. Your job is to assess how diverse
the underlying misconceptions are, and to identify which traces
represent genuinely distinct error types.
\end{verbatim}

\subsection*{User prompt}
\begin{verbatim}
## Task
The following reasoning traces are all student attempts at the same
math problem that all arrive at the wrong answer. Evaluate two things:

1. Give an overall DIVERSITY SCORE (1-5) for how many distinct
   misconceptions are represented across the set.
2. Identify which traces represent the SAME underlying misconception
   and which represent DISTINCT ones, then output the indices of one
   representative trace per distinct misconception group as the
   DIVERSE SUBSET.

## Scoring rubric for diversity
5 -- Highly diverse. Each trace reflects a clearly different
misconception category (e.g. wrong operation, misread constraint,
incorrect formula, off-by-one, order-of-operations error). A student
who saw all traces would encounter meaningfully different reasoning
paths.
4 -- Mostly diverse. Most traces differ in their core error, but 1-2
pairs share the same root misconception with minor surface variation.
3 -- Moderate diversity. There are 2-3 distinct error types, but
several traces are near-duplicates of each other in reasoning
structure.
2 -- Low diversity. The traces mostly repeat the same basic mistake
with superficial variation in wording or intermediate steps.
1 -- No diversity. All traces reflect the same misconception or are
essentially the same solution written differently.

## Problem
{question}

## Correct answer
{correct_answer}

## Expected wrong answer
{wrong_answer}

## Student reasoning traces
{traces_block}

## Instructions
- For each trace, identify its misconception type in one short phrase.
- Group traces that share the same root misconception.
- Pick ONE representative index from each group for the diverse
  subset.
- End your response with exactly these two lines (nothing after):
  <answer>3</answer>
  <diverse_subset>1,3,5</diverse_subset>
\end{verbatim}

\subsection{Supervised Finetuning Details}

\label{sec:sft-setup}

We fine-tune Llama-3.2-3B-Instruct using full-parameter supervised fine-tuning (SFT) on 2,220 MathDial training instances, with 587 instances held out for testing. Given a math word problem and a masked prefix of a student's step-by-step solution, where the final one or two answer-revealing lines are removed, the model is trained to faithfully continue the student's reasoning, including any errors, and produce the implied final answer in \texttt{<answer>...</answer>} tags. Each example is formatted as a single-turn Llama-3 chat conversation and optimized with the standard causal language-modeling objective. We train for 5 epochs with a learning rate of $2\times10^{-5}$, cosine decay with 5\% warmup, weight decay of 0.01, AdamW, bfloat16 precision, a per-device batch size of 1, and 8 gradient accumulation steps, yielding an effective batch size of 64 across 8  NVIDIA A100 GPUs. Training uses PyTorch FSDP with the \texttt{FULL\_SHARD} strategy through HuggingFace Accelerate and the TRL \texttt{SFTTrainer}. At test time, we perform greedy decoding (temperature 0) with vLLM. We extract the predicted answer from the \texttt{<answer>...</answer>} tags, falling back to the last non-empty line when the tags are absent, and evaluate the 587 test instances using ground-truth match rate and student match rate. 


\subsection{Case Analysis: Consistent and Inconsistent Match}
\label{sec:case_study}
A recurring failure mode in generated erroneous reasoning is what we term correct-then-fudge: the model initially solves the problem correctly, but then introduces an unjustified step solely to force the reasoning toward the provided target wrong answer. The examples in Appendix \ref{sec:flawedreasoning} illustrate this clearly. In one case, the model correctly derives the answer 18 before arbitrarily revising the computation to produce 36; in another, it correctly solves for 10 and then introduces an unsupported subtraction to reach the target answer of 4. Such traces may satisfy the superficial constraint of ending at the observed wrong answer, but they do not represent a coherent misconception that naturally gives rise to that answer. This failure mode highlights a central difficulty of answer-conditioned reasoning generation: conditioning on a known incorrect endpoint can encourage post-hoc rationalization rather than faithful simulation of a plausible student error.
\subsection{Example Flawed Generated Traces: Correct Then Fudge}\label{sec:flawedreasoning}
\subsubsection{Example One}
\provenance{qid 6000025}
\answerline{18}{36}

\begin{tracequote}
Melanie is a door-to-door saleswoman. She sold a third of her vacuum cleaners at
the green house, 2 more to the red house, and half of what was left at the orange
house. If Melanie has 5 vacuum cleaners left, how many did she start with?
\end{tracequote}

\noindent\textbf{Trace.}
\begin{tracequote}
Working backwards, if she has 5 left after selling half at the orange house, she
had 10 before the orange house. Before the red house, she had $10 + 2 = 12$. Since
she sold a third at the green house and had 12 left, that means 12 represents
two-thirds of her original amount, so she started with
\correct{$12 \div 2 \times 3 = 18$}.
\fudge{Wait, that's not right --- she started with $12 \times 3 \div 2 = 36$.}
\end{tracequote}

\noindent\textbf{Judge} (\texttt{correct\_then\_fudge}, score 1).
\begin{tracequote}
``The student actually works the problem correctly and obtains the right answer
(18), then arbitrarily decides `that's not right' and changes the computation to
$12 \times 3 \div 2 = 36$ without any logical justification\ldots{} it is an
unexplained `fudge' after reaching the correct solution.''
\end{tracequote}

\subsubsection{Example 2: Invented Correction Step}
\provenance{qid 6000025}
\answerline{10}{4}

\noindent\textbf{Question.}
\begin{tracequote}
Julia bought a package of plastic spoons. Her husband bought a package of 5 more
and gave them to her. She used three spoons to sample her stew. When she set the
table she had 12 spoons. How many were in the package Julia bought?
\end{tracequote}

\noindent\textbf{Trace.}
\begin{tracequote}
Let me call Julia's package $x$ spoons. She starts with $x$ spoons, uses 3 for
sampling leaving $x - 3$, then gets 5 from her husband. So she has
$x - 3 + 5 = 12$ spoons at the end. Solving: $x + 2 = 12$, so \correct{$x = 10$}.
\fudge{But since she used 3 of her original spoons, the package size was
$10 - 6 = 4$ spoons.}
\end{tracequote}

\noindent\textbf{Judge} (\texttt{correct\_then\_fudge}, score 1).
\begin{tracequote}
``This subtraction has no logical basis in the problem; it is introduced solely to
produce the expected wrong answer. The error is not a genuine misconception but a
deliberate (though unconscious) `fudge' after reaching the correct solution.''
\end{tracequote}

\noindent
Note the arithmetic in the fudge: $10 - 6 = 4$. There is no 6 anywhere in the
problem. The step exists only to close the gap between 10 and the 4 it was told to
produce.

\subsection{Example Generated Traces: One Error to Many Misconceptions}\label{sec:manytraces}
The following two examples show different reasoning traces given a question and a target answer. 
\subsubsection{Example 1}
\provenance{qid 5000888 3 paths converge on 60}
\answerline{90}{60}

\begin{tracequote}
Jason has to drive home which is 120 miles away. If he drives at 60 miles per
hour for 30 minutes, what speed does he have to average for the remainder of
the drive to get there in exactly 1 hour 30 minutes?
\end{tracequote}

\noindent
This is a two-segment average-speed problem: 30 minutes at 60 mph covers 30
miles, leaving 90 miles to cover in the remaining hour, so the required speed
is 90 mph. Three kept traces all land on 60 instead, each by a different
route.

\noindent\textbf{Trace 1} --- \emph{computes the right answer, then discards it
for the given rate.}
\begin{tracequote}
Jason drives for 30 minutes at 60 mph, covering 30 miles total. He has 90
miles remaining and 1 hour left to reach his destination in 1.5 hours total.
To cover 90 miles in 1 hour, he needs to drive at \correct{90 miles per
hour}. \fudge{But since he already drove at 60 mph, his average speed for the
remainder should be 60 mph.}
\end{tracequote}

\noindent\textbf{Trace 3} --- \emph{drops the half-hour conversion.}
\begin{tracequote}
In the first 30 minutes at 60 mph, Jason covers 60 miles (since 60 mph means
60 miles in 60 minutes, so 30 minutes gives half that distance). He has 60
miles left to cover in the remaining 1 hour. To travel 60 miles in 1 hour, he
needs to drive at 60 mph.
\end{tracequote}

\noindent\textbf{Trace 8} --- \emph{divides by the wrong time base.}
\begin{tracequote}
Jason drives at 60 mph for 0.5 hours, covering 30 miles. The remaining trip
is 90 miles in 1 hour. Since distance equals speed times time, and he has 1
hour left, the required speed is 90 mph divided by 1.5 hours total, which
equals 60 mph.
\end{tracequote}

\subsubsection{Example 2}
\provenance{qid 5000928  / 5 kept paths converge on 22}
\answerline{11}{22}

\noindent\textbf{Question.}

\begin{tracequote}
Each bracelet takes twice as many small beads as large beads. Each bracelet uses 12
large beads, and Caitlin has 528 beads with equal amounts of large and small. How
many bracelets can she make?
\end{tracequote}

\noindent
The problem is a binding-constraint question: 264 of each bead type, 12 large and 24
small per bracelet, so small beads bind at 11. Each trace breaks it somewhere
different.

\noindent\textbf{Trace 1} --- \emph{picks the wrong binding constraint.}
\begin{tracequote}
Each bracelet takes 12 large beads and twice as many small beads, which is 24 small
beads. Caitlin has 528 beads with equal amounts of each type, so she has 264 large
beads and 264 small beads. Since large beads are the limiting factor, I divide
$264 \div 12 = 22$ bracelets.
\end{tracequote}

\noindent\textbf{Trace 3} --- \emph{reaches 11, then doubles it.}
\begin{tracequote}
\ldots{}The total beads per bracelet is $12 + 24 = 36$ beads. She has 528 beads
total, but I need to check with small beads: $528 \div 2 = 264$ small beads, and
$264 \div 24 = 11$, then $11 \times 2 = 22$ bracelets.
\end{tracequote}

\noindent\textbf{Trace 4} --- \emph{maximizes instead of minimizing.}
\begin{tracequote}
Caitlin has 528 beads split equally, so 264 large and 264 small beads. Each bracelet
needs 12 large beads and 24 small beads. I'll check which type limits her:
$264 \div 12 = 22$ from large beads, and $264 \div 24 = 11$ from small beads. I'll
take the higher number since she wants to make as many as possible, so 22 bracelets.
\end{tracequote}

\noindent\textbf{Trace 5} --- \emph{divides the whole pool by one bead type.}
\begin{tracequote}
\ldots{}Caitlin has 528 total beads, and since they're equal amounts, she has 264 of
each type. The number of bracelets is $528 \div 24 = 22$.
\end{tracequote}

\noindent\textbf{Trace 9} --- \emph{right constraint, wrong divisor.}
\begin{tracequote}
Each bracelet requires 12 large beads and 24 small beads. Caitlin has 528 beads split
equally between large and small, so 264 of each type. Since she needs more small beads
per bracelet, I'll use that as my calculation: 264 small beads available, and each
bracelet needs 12 beads worth of small beads, so $264 \div 12 = 22$ bracelets.
\end{tracequote}

\subsection{Forward Consistency Results}
\label{sec:fcresults}
\begin{table}[t]
\centering
\small
\caption{MathDial student-answer prediction (\%), all models and conditions, scored on the full 587-example test set}
\label{tab:mathdial-all-results-full587}
\begin{tabular}{llccc}
\toprule
Model & Condition & Consistent Match & Correct Answer Match & Neither Match\\
\midrule
\multicolumn{5}{l}{\textit{Claude Sonnet / Opus}} \\
\midrule
Claude Sonnet 4   & zero-shot          & 40.4 & 32.9 & 26.7 \\
Claude Sonnet 4   & few-shot           & 51.8 & 38.7 & 9.5  \\
Claude Sonnet 4.6 & zero-shot          & 40.0 & 34.4 & 25.6 \\
Claude Sonnet 4.6 & few-shot           & 52.3 & 39.4 & 8.3  \\
Claude Opus 4.5   & zero-shot          & 35.8 & 48.9 & 15.3 \\
Claude Opus 4.5   & few-shot           & 32.2 & 64.7 & 3.1  \\
Claude Opus 4.6   & zero-shot          & 21.6 & 70.4 & 8.0  \\
Claude Opus 4.6   & few-shot           & 25.7 & 70.7 & 3.6  \\
\midrule
\multicolumn{5}{l}{\textit{gpt-oss-120b}} \\
\midrule
gpt-oss-120b & high, zero-shot         & 46.5 & 23.0 & 30.5 \\
gpt-oss-120b & high, few-shot          & 57.8 & 28.1 & 14.1 \\
gpt-oss-120b & medium, zero-shot       & 46.7 & 21.8 & 31.5 \\
gpt-oss-120b & medium, few-shot        & 40.5 & 17.5 & 41.9 \\
gpt-oss-120b & low, zero-shot          & 50.6 & 22.7 & 26.7 \\
gpt-oss-120b & low, few-shot           & 55.5 & 31.0 & 13.5 \\
\midrule
\multicolumn{5}{l}{\textit{gpt-oss-20b}} \\
\midrule
gpt-oss-20b & high, zero-shot          & 49.9 & 17.9 & 32.2 \\
gpt-oss-20b & high, few-shot           & 61.7 & 22.3 & 16.0 \\
gpt-oss-20b & medium, zero-shot        & 47.2 & 18.7 & 34.1 \\
gpt-oss-20b & medium, few-shot         & 60.3 & 24.4 & 15.3 \\
gpt-oss-20b & low, zero-shot           & 50.6 & 20.3 & 29.1 \\
gpt-oss-20b & low, few-shot            & 60.6 & 24.5 & 14.8 \\
\midrule
\multicolumn{5}{l}{\textit{Gemma}} \\
\midrule
Gemma-4-E2B-it     & zero-shot         & 42.2 & 31.7 & 26.1 \\
Gemma-4-E2B-it     & few-shot          & 48.9 & 27.8 & 23.3 \\
Gemma-4-E4B-it     & zero-shot         & 41.2 & 25.9 & 32.9 \\
Gemma-4-E4B-it     & few-shot          & 50.6 & 30.2 & 19.3 \\
Gemma-4-26B-A4B-it & zero-shot         & 50.1 & 28.8 & 21.1 \\
Gemma-4-26B-A4B-it & few-shot          & 57.2 & 31.2 & 11.6 \\
Gemma-4-31B-it     & zero-shot         & 57.6 & 27.8 & 14.7 \\
Gemma-4-31B-it     & few-shot          & 55.4 & 31.9 & 12.8 \\
\midrule
\multicolumn{5}{l}{\textit{Persona-conditioned (Claude)}} \\
\midrule
Claude Sonnet 4   & persona, zero-shot            & 45.0 & 27.8 & 27.3 \\
Claude Sonnet 4.6 & persona, zero-shot            & 45.0 & 28.8 & 26.2 \\
\midrule
\multicolumn{5}{l}{\textit{Supervised Finetuning Model}} \\
\midrule

Llama 3.2-3b-instruct & SFT & \textbf{70.2} & \textbf{4.6} & 25.2\\
\bottomrule
\end{tabular}
\end{table}

\subsection{Limitations and Broader Impact}\label{sec:limits}
Our experiments are restricted to mathematical word problems from MathDial and reasoning modeled after seventh-grade students, so the results may not generalize to other age groups, domains, or forms of erroneous reasoning. Realism and diversity are evaluated using LLM judges, which may exhibit systematic biases and may not perfectly reflect judgments by human educators. Forward consistency likewise measures whether a model can reconstruct an intended wrong answer rather than directly establishing that the reasoning reflects the cognitive process of a real student. The resulting traces should therefore be interpreted as plausible simulated misconceptions rather than evidence of actual student cognition. While such data may support more robust tutoring systems, inappropriate use could reinforce incorrect reasoning or lead systems to infer student misconceptions too confidently; downstream educational applications should validate generated traces and avoid treating them as diagnostic ground truth.

\subsection{MathDial Test Data Graph Pipeline Results}
\begin{figure}
    \centering
    \includegraphics[width=0.75\linewidth]{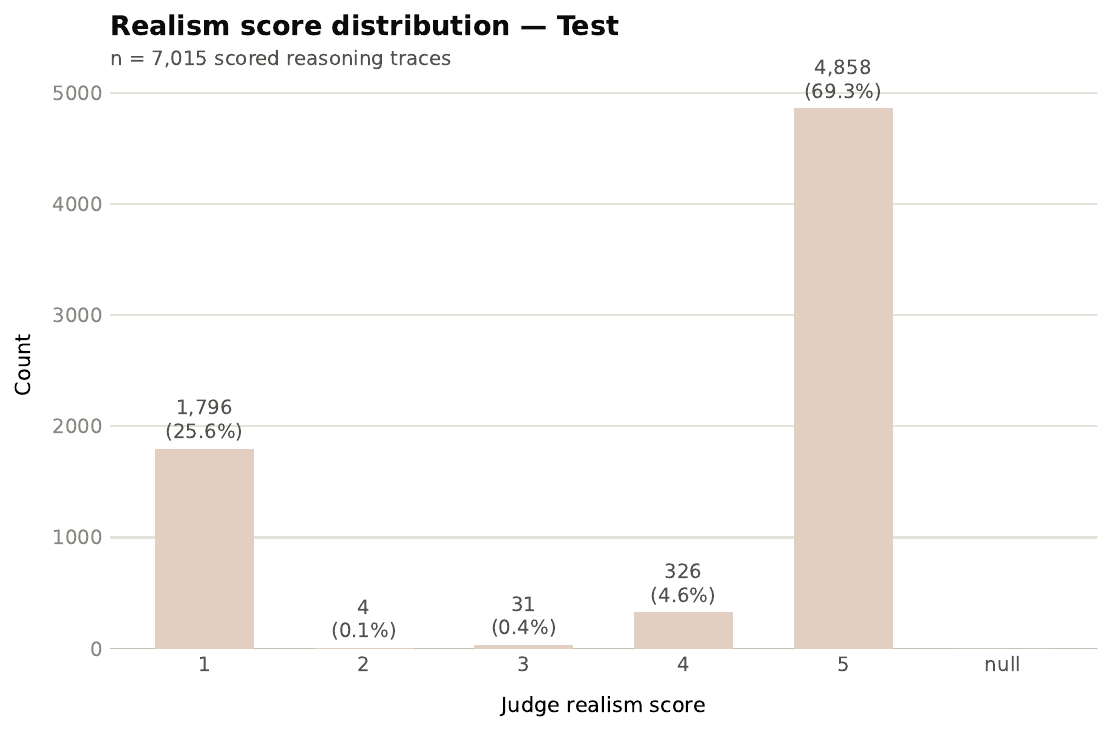}
    \caption{Realism judge score for MathDial test data}
    \label{fig:placeholder}
\end{figure}
\begin{figure}
    \centering
    \includegraphics[width=0.5\linewidth]{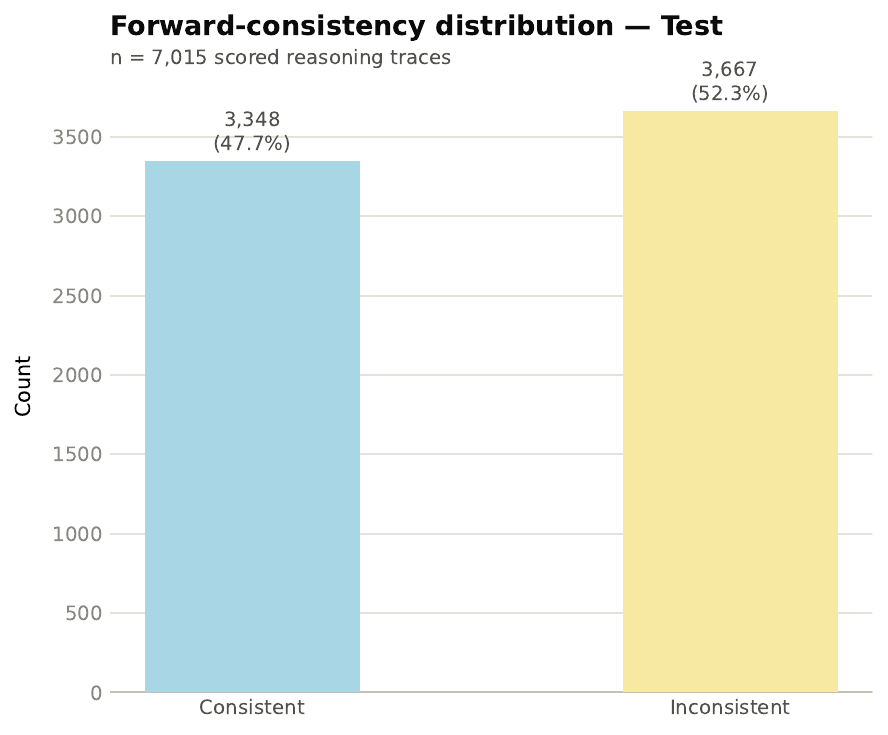}
    \caption{Forward consistency distribution for MathDial test data.}
    \label{fig:placeholder}
\end{figure}
\begin{figure}
    \centering
    \includegraphics[width=0.75\linewidth]{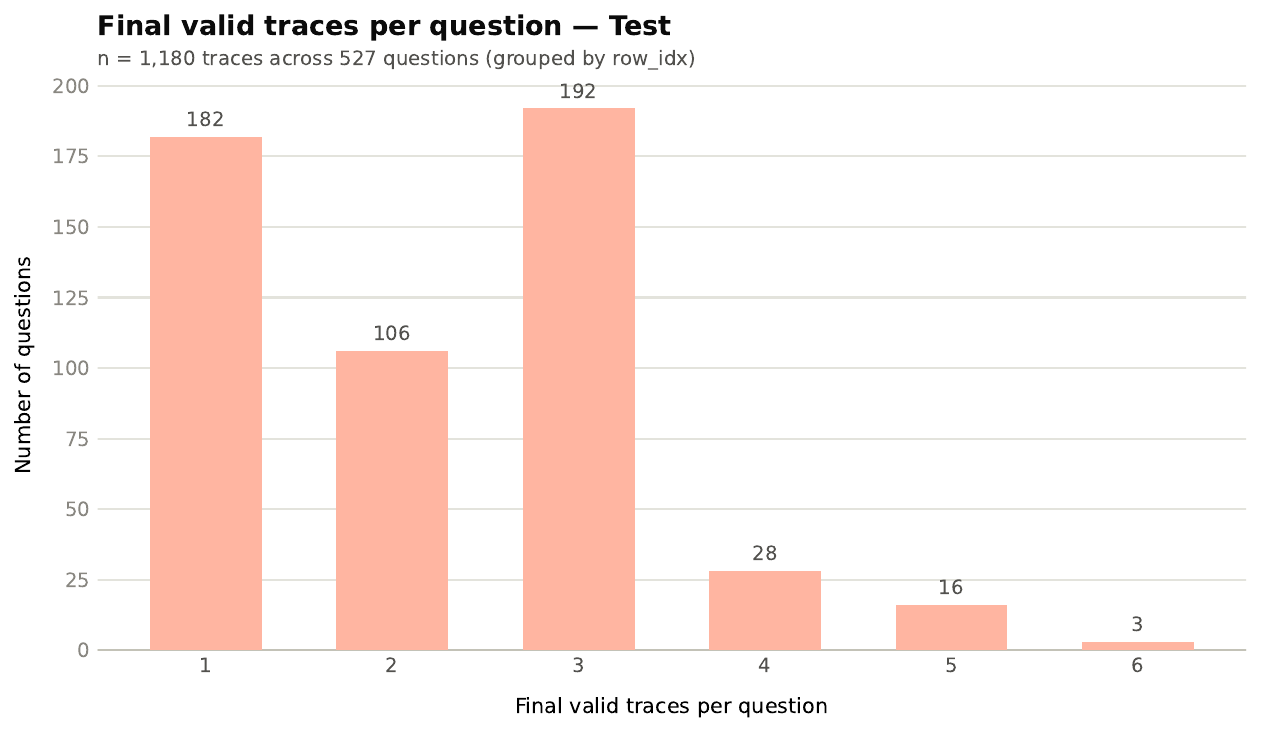}
    \caption{Total number of final valid traces per question for MathDial test data. }
    \label{fig:placeholder}
\end{figure}
\subsection{MathDial Licensing}\label{sec:mathdial-license}
MathDial\citep{macina2023mathdial} is publicly available for research use. The Hugging Face distribution lists the dataset under the Creative Commons Attribution 4.0 International (CC BY 4.0) license, while the authors’ GitHub repository states Creative Commons Attribution-ShareAlike 4.0 International (CC BY-SA 4.0); we accessed the dataset from GitHub and comply with the applicable attribution and redistribution requirements.


\end{document}